\documentclass[conference]{IEEEtran}

\usepackage{amsmath}
\usepackage{graphicx}
\usepackage{multirow}
\usepackage{array}
\usepackage{booktabs}
\usepackage[left=54pt, right=54pt, bottom=54pt, top=54pt]{geometry}

\usepackage{url}

\usepackage{hyperref}
\usepackage{cleveref}
\usepackage{etoolbox}
\usepackage{pifont} 
\usepackage{xcolor} 
\usepackage{graphicx} 
\usepackage{makecell}
\usepackage[nolist,nohyperlinks]{acronym}

\definecolor{heavygreen}{RGB}{0,128,0} 

\newcommand{\cmark}{\textcolor{heavygreen}{\scalebox{1.5}{\ding{51}}}} 
\newcommand{\xmark}{\textcolor{red}{\scalebox{1.5}{\ding{55}}}} 

\makeatletter
\patchcmd{\@makecaption}
  {\scshape}
  {}
  {}
  {}
\makeatother

\begin{document}
%
\title{Explainable Lane Change Prediction for Near-Crash Scenarios Using Knowledge Graph Embeddings and Retrieval Augmented Generation}
%
%
%

\author{M. Manzour$^{1}$, A. Ballardini$^{1}$, R. Izquierdo$^{1}$, and M. A. Sotelo$^{1}$\\
$^{1}$Department of Computer Engineering, University of Alcal\'a, Madrid, Spain\\
$[$ahmed.manzour, augusto.ballardini, ruben.izquierdo, miguel.sotelo$]$@uah.es}

\begin{acronym}
\acro{RAG}{Retrieval Augmented Generation}
\acro{KG}{Knowledge Graph}
\acro{SVM}{Support Vector Machine}
\acro{ANN}{Artificial Neural Network}
\acro{NGSIM}{Next Generation Simulation}
\acro{RNN}{Recurrent Neural Network}
\acro{LSTM}{Long Short-Term Memory}
\acro{MHI}{Motion History Image}
\acro{CNN}{Convolutional Neural Network}
\acro{XGBoost}{eXtreme Gradient Boosting}
\acro{KGE}{Knowledge Graph Embedding}
\acro{TTC}{Time To Collision}

\acro{CRASH}{CARLA Risky-lane-change Anticipation in Simulated Highways}
\acro{NHTSA}{National Highway Traffic Safety Administration}
\acro{AV}{Autonomous Vehicle}
\acro{LLM}{Large Language Model}
\acro{MRR}{Mean Reciprocal Rank}
\acro{IDM}{Intelligent Driver Model}

\end{acronym}

\maketitle

\begin{abstract}
Lane-changing maneuvers, particularly those executed abruptly or in risky situations, are a significant cause of road traffic accidents. However, current research mainly focuses on predicting safe lane changes. Furthermore, existing accident datasets are often based on images only and lack comprehensive sensory data. In this work, we focus on predicting risky lane changes using the \ac{CRASH} dataset (our own collected dataset specifically for risky lane changes), and safe lane changes (using the HighD dataset). Then, we leverage \acp{KG} and Bayesian inference to predict these maneuvers using linguistic contextual information, enhancing the model's interpretability and transparency. The model achieved a $91.5\%$ f1-score with anticipation time extending to four seconds for risky lane changes, and a $90.0\%$ f1-score for predicting safe lane changes with the same anticipation time. We validate our model by integrating it into a vehicle within the CARLA simulator in scenarios that involve risky lane changes. The model managed to anticipate sudden lane changes, thus providing automated vehicles with further time to plan and execute appropriate safe reactions. Finally, to enhance the explainability of our model, we utilize \ac{RAG} to provide clear and natural language explanations for the given prediction.
\end{abstract}

\IEEEpeerreviewmaketitle

\section{Introduction}
\begin{figure}[ht]
\centering
\includegraphics[width=\columnwidth]{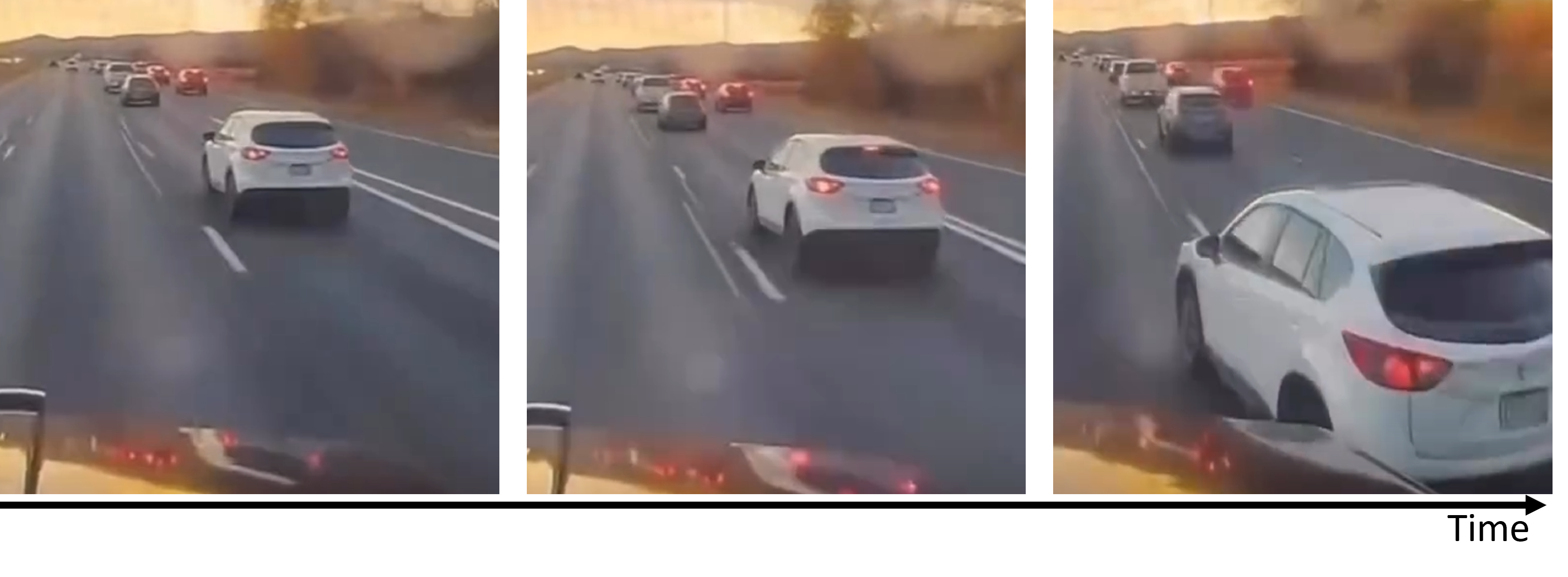}
\vspace{-9mm}
\caption{Target vehicle made a risky left lane change that led to a collision.}
\label{fig:introduction_lane_changing_example}
\end{figure}

\IEEEPARstart{L}{ane}-changing maneuvers (especially abrupt lane changes) are one of the causes of vehicle crashes, as a report indicated that $33\%$ of all road crashes occur due to driver decision errors including lane changes \cite{nhtsa2015critical}. A risky lane change occurs when drivers switch lanes suddenly without adequate warning, which leaves insufficient time for other vehicles to react, leading to frequent near-crashes or collisions. In contrast, a safe lane change usually comes with early signals or clear intentions to change lanes, giving other drivers sufficient time to react. While recent research has mainly focused on predicting safe lane changes, there is a critical gap when it comes to addressing risky lane-changing scenarios. \Cref{fig:introduction_lane_changing_example} shows a risky lane change in which the white vehicle was involved in a risky situation, it was forced to change lanes to the left with little warning which led to a severe collision. Ignoring these unsafe lane-changing events makes it difficult for automated systems to anticipate and prevent near-crash situations. Moreover, many models that use numerical inputs function like black boxes, making it tough to interpret or explain their outputs. This lack of transparency poses challenges when trying to justify predictions to users who may not be familiar with the underlying algorithms. Additionally, existing accident datasets do not provide sufficient input for designing explainable prediction models. These datasets typically consist of video recordings without accompanying numerical or linguistic data, such as the velocities or positions of vehicles. This absence of crucial information makes it hard to develop models that provide clear and comprehensive predictions.
To address these challenges, this work focuses on addressing the following points:
\begin{enumerate}
    \item A customized dataset (\ac{CRASH} dataset) of risky lane-changing scenarios is developed using the CARLA simulator, reconstructing real-world scenes with detailed numerical data.
    \item An interpretable and transparent prediction model for safe and risky lane changes is developed by combining \acp{KG} with Bayesian inference and incorporating contextual linguistic information to improve interpretability.
    \item The model is validated in the CARLA simulator, demonstrating that it can predict unsafe maneuvers in advance and provide the vehicle with sufficient time to react accordingly.
    \item The model's explainability is increased by using \ac{RAG} to provide clear natural language explanations for the predictions.
\end{enumerate}

The rest of this article is organized as follows. \Cref{sec:sota} presents the state of the art. \Cref{sec:dataset} contains a brief introduction to the \ac{CRASH} dataset. Then, our proposed methodology is discussed in detail in \cref{sec:methodology}. In \cref{sec:results}, results will be presented. Finally, \cref{sec:concliusions} concludes the work and provides some recommendations for future work.

\section{State of the Art}\label{sec:sota}
Recently, different works have focused on predicting vehicle lane changes using different types of inputs and using different models.
In 2019, \cite{benterki2019prediction} used two machine learning models; \ac{SVM} and \ac{ANN} to predict lane changes of surrounding vehicles on highways. They used the \ac{NGSIM} dataset, their input features included vehicle's speeds and accelerations in both longitudinal and lateral directions, distances to adjacent lane boundaries, yaw angles and yaw rates. Also in 2019, \cite{izquierdo2019experimental} focused on predicting the lane-changing intentions of surrounding vehicles using only visual information from the PREVENTION dataset. They explored two methods: the first was a \ac{MHI} combined with a \ac{CNN}, where temporal and visual data were fed into the \ac{CNN}. The second method involved a GoogleNet-LSTM model, where features extracted by a GoogleNet \ac{CNN} were passed to an \ac{LSTM} to learn temporal patterns. Inputs included RGB images, the center coordinates \textit{(x,y)}, and the dimensions of the bounding boxes around vehicles.
In 2020, \cite{laimona2020implementation} trained \ac{RNN} and \ac{LSTM} models using the PREVENTION dataset to predict lane-changing intentions by tracking the positions of surrounding vehicles, specifically the centers of their bounding boxes. In 2018, \cite{su2018learning} employed an \ac{LSTM} model to predict lane changes by considering both the vehicle's past trajectory and the states of surrounding vehicles. Using data from the \ac{NGSIM} dataset, they incorporated features such as the vehicle’s lateral and longitudinal positions relative to the lane, its acceleration, the presence of vehicles to the right or left, and the distances to surrounding vehicles. In 2022, \cite{xue2022integrated} utilized \ac{XGBoost} and \ac{LSTM} models to predict lane change decisions and future trajectories using scenarios from the HighD dataset. Their approach considered factors like traffic density, vehicle type, and the relative movements between the target vehicle and the surrounding vehicles. Initially, they built a traffic flow model using the target vehicle's longitudinal speed and acceleration, headway distance, and relative velocities to the surrounding vehicles. In 2023, a dual transformer model was developed by \cite{gao2023dual}, consisting of two parts: one for predicting lane changes and another for forecasting trajectories. The first transformer used the target vehicle's historical lateral movements and the states of surrounding vehicles, including distances and velocities relative to the target vehicle. The output from this lane change prediction was then combined with the target vehicle's past lateral movements and fed into the second transformer to link intentions with future trajectories. This model was trained and tested on the HighD and \ac{NGSIM} datasets. In 2024, \cite{manzour2024vehicle} used \ac{KGE} followed by Bayesian inference which acted as a downstream task on the grounds of the learned embedding to predict safe lane changes on the HighD dataset. They considered inputs in linguistic formats to enhance the model's interpretability. The considered features are the vehicle lateral velocity and acceleration, \ac{TTC} risk with the surrounding vehicles. Work \cite{hussien2024rag} used the same methodology. However, it extends the architecture by adding the RAG mechanism which leverages the power of \acp{LLM} to generate context-specific explanations that are grounded in external, verifiable knowledge, and provide explanations of the scene provided with the reason for the lane change in natural language which is easy to understand by end-users. The work in \cite{peng2024lc} addressed predicting safe lane changes and trajectories in the HighD dataset by leveraging the strong reasoning capabilities and self-explanation abilities of \acp{LLM}. Data is processed as numerical values in natural language prompts to be the input to the \acp{LLM}. Then, they utilized Chain-of-Thought reasoning to enhance prediction explainability. The advantage of this work is that it achieves high-performance metrics and fine natural language explanations to the end users with only the powers of the tuned \acp{LLM}. The challenge is that even though the \acp{LLM} provides explanations and employs chain-of-thought reasoning to interpret input prompts, there remains a gap in the transparency and grounding of the LLM reasoning process as these models rely heavily on internal representations (the knowledge and patterns that the \acp{LLM} has learned internally during its training). This reliance can make the \acp{LLM} reasoning process less transparent, and the explanations might be difficult to verify or to be traced back to the inputs. So, the way it processes information internally is not easily visible or understandable to the users. Also, users and experts still cannot trace the prediction process starting from having the numerical input data till the prediction, they can't see how a change in the input can affect the probability of the final prediction, and it is often considered a ``black box" in that case. On the other side, integrating \acp{KG} and Bayesian inference improves interpretability and transparency by creating a structured and interpretable framework where users can trace the process of obtaining the prediction and can show the effect of each input on the probability of the final prediction through the priors given by the KG for every event (input), which is something hard to trace in the tuned LLM approach. Moreover, without grounding in external knowledge, \acp{LLM} explanations may be less reliable. However, by using RAG, we ensure that explanations are grounded in external, verifiable knowledge, enhancing reliability. Finally, \Cref{tab:SOTA} presents a comparison of the literature discussed previously to highlight the identified research gaps. The first column indicates whether the models in the related works use numerical inputs or linguistic explainable inputs. The second and third columns specify whether the work addresses safe or risky lane changes respectively. The fourth column notes whether the related work includes experimental validation for simple decision-making, even if conducted in simulation. The fifth column indicates whether the architecture provides a natural language explanation. The last row summarizes the aspects that our work addresses.
\begin{table*}[ht]
\caption{Comparison of related works on lane change prediction based on input types, focus, and explainability.}
\vspace{-5mm}
\begin{center}
\begin{tabular}{|c|c|c|c|c|c|}
\hline
      Work & \makecell{Linguistic Explainable Inputs}  & Safe Lane Changes & Risky Lane Changes & Decision Making & \makecell{Natural Language Explanation}\\
\hline
\makecell{\cite{benterki2019prediction, izquierdo2019experimental, laimona2020implementation}\\ \cite{su2018learning, xue2022integrated, gao2023dual}}  & \xmark & \cmark & \xmark & \xmark & \xmark\\
\hline
\cite{manzour2024vehicle}   & \cmark & \cmark & \xmark & \xmark & \xmark\\
\hline
\cite{hussien2024rag} & \cmark & \cmark & \xmark & \xmark & \cmark\\
\hline
\cite{peng2024lc} & \xmark & \cmark & \xmark & \xmark & \cmark\\
\hline
Ours & \cmark & \cmark & \cmark & \cmark & \cmark\\
\hline
\end{tabular}
\label{tab:SOTA}
\end{center}
\vspace{-8mm}
\end{table*}
After a thorough examination of the existing literature, we have identified the following research gaps, which are addressed in this work:
\begin{enumerate}
    \item \textbf{Specialized dataset for risky lane changes:} There is a lack of datasets focusing specifically on risky lane-change scenarios. We address this gap by collecting the \ac{CRASH} dataset using the CARLA simulator.
    
    \item \textbf{Prediction based on contextual linguistic information:} Current models rely on trajectory and numerical data from past experiences. We propose a method to predict both safe and risky lane-change maneuvers using contextual linguistic information, moving beyond mere numerical inputs. 
    
    \item \textbf{Interpretable and transparent predictions:} Many existing models lack interpretability and transparency. Our approach utilizes \ac{KGE} and Bayesian inference to provide predictions that are both interpretable and transparent.

    \item \textbf{Validation in simulation environment:} There is a need for models to be validated in realistic simulation settings. We validate our model within the CARLA simulator, demonstrating its ability to predict unsafe maneuvers in advance and provide the vehicle with sufficient time to respond appropriately.

    \item \textbf{Enhanced explainability through RAG integration:} To further improve the model's reliability and explainability, we integrate RAG which provides clear, natural language explanations for the predictions based on external, reliable knowledge sources. This integration enriches the architecture's explainability, making the target vehicle's decision-making process more understandable to users.
\end{enumerate} 

Addressing these gaps will not only give the ability to provide explanations but also make the architecture more suitable for safety-critical applications where input-to-output traceability, understanding, and transparency are essential.

\section{Dataset}\label{sec:dataset}
This section briefly introduces the \ac{CRASH} dataset which focuses on risky lane-changing maneuvers. The dataset is created by examining videos from various datasets containing different real accident and near-crash scenarios, some of these scenarios include lane changes \cite{yao2022dota}\cite{kezebou2022highway}. These datasets were collected from front-view cameras mounted on vehicles. However, they only provide images and lack numerical data about the vehicle's state and the state of the surrounding vehicles. Therefore, CARLA simulator is utilized to recreate similar scenes \cite{Dosovitskiy17}. By integrating CARLA, it is possible to generate comparable scenarios and obtain numerical values for all required inputs concerning the ego vehicle, target vehicle (which does the risky maneuver), and all surrounding vehicles in the scene. The dataset contains detailed information about every vehicle in the scene. For each vehicle, data are provided on its \textit{(x, y)} location within the map, velocities, and accelerations in the lateral and longitudinal directions, braking status (whether the vehicle is braking or not), rear braking light and turning signal status, lane position, position relative to the center of the lane, gap distance and time to collision with the surrounding vehicles. Additionally, if there is an adjacent left or right vehicle, the lateral distance between the vehicle and the adjacent vehicle is included. The dataset also includes lane densities for the left, center, and right lanes, all measured relative to the vehicle’s position, as well as the average speeds of these lanes. Furthermore, the lane attraction score is provided, which is calculated as each lane's average speed divided by its density, it gives an intuition about which lane will be more attractive to the vehicle based on the average velocity and density of that lane. The dataset contains a total of $50$ samples, including $25$ left lane changes, and $25$ right lane changes, all these samples are classified as risky lane-changing maneuvers that end up with a collision or a near-crash. Safe maneuvers are not included because existing datasets, such as the HighD dataset, already cover safe lane-changing maneuvers. The scenarios are recorded from two different perspectives. The first is the front view from the ego vehicle's perspective, providing a realistic view of what a driver would see during the accident. The second is the bird's eye view, which allows the observer to see the entire situation, including how the ego vehicle interacts with the target vehicle's risky maneuver and all surrounding vehicles.
\section{Methodology}\label{sec:methodology}
\subsection{Architecture Overview} \label{subsec:arquitecture}
\begin{figure*}[ht]
\centering
\includegraphics[width=\linewidth]{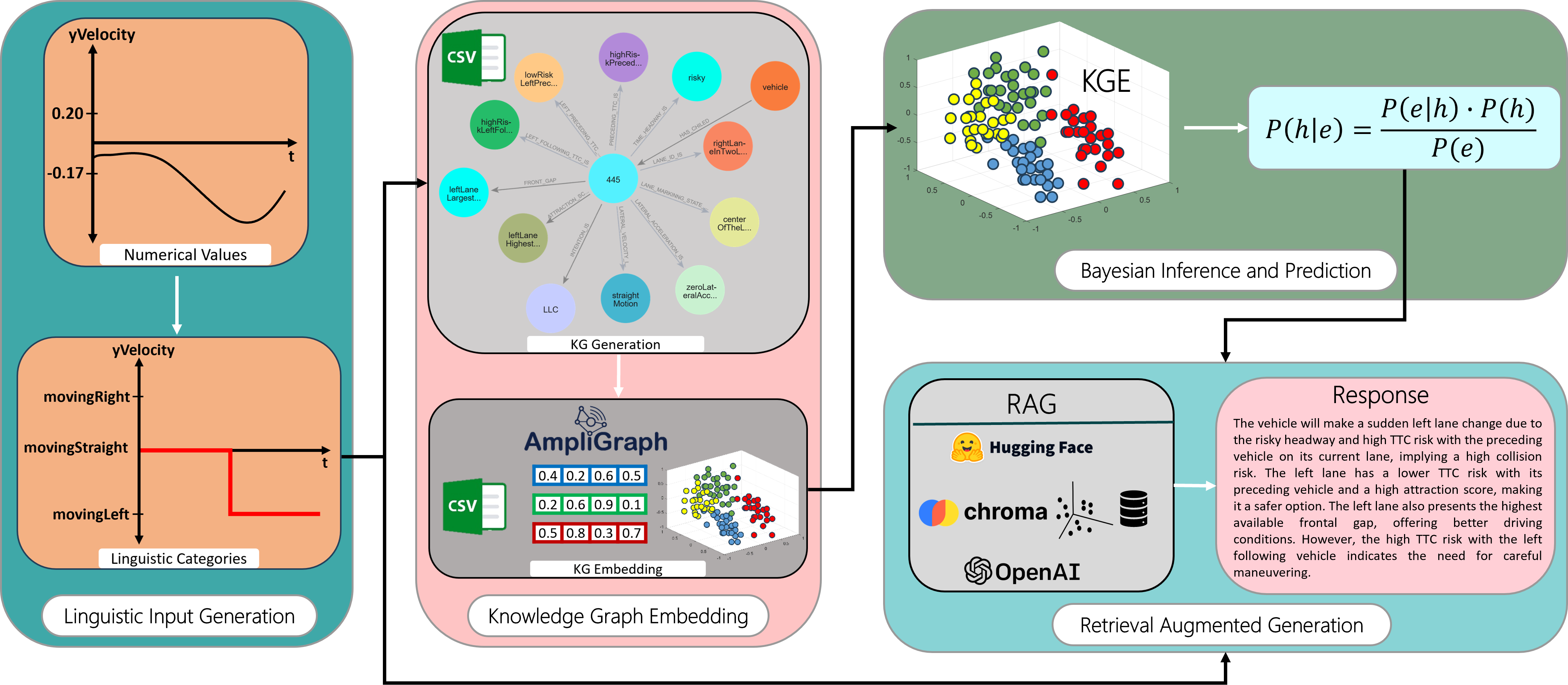}
\vspace{-6mm}
\caption{The pipeline for anticipating and explaining lane changes consists of four phases: Linguistic Input Generation, Knowledge Graph Embedding, Bayesian
Inference and Prediction, and Retrieval Augmented Generation.}
\label{fig:methodology}
\vspace{-5mm}
\end{figure*}

Our architecture is divided into four stages as indicated in \cref{fig:methodology}.
The first stage is Linguistic Input Generation, where we take our numerical input values and convert it to linguistic categories. This is obtained based on literature and the data distribution along the lane-changing categories. For the second stage (Knowledge Graph Embedding), we use the linguistic inputs to create our \ac{KG}, which is in the form of triples inside a CSV file. Then, this \ac{KG} is embedded using the Ampligraph library. In the third stage, we carry out Bayesian inference on the grounds of these learned embeddings to get our final prediction. Finally, we feed this prediction along with the linguistic inputs to the Retrieval Augmented Generation stage in order to get an explanation of these predictions and the target vehicle decision-making process.
\subsection{Input and Knowledge Graph Ontology Definition}
From the dataset, the features selected include the y-velocity and y-acceleration of the target vehicle, the vehicle's lane position, distance from the center of the vehicle to the center of the lane, the time headway between the target vehicle and the preceding vehicle, time to collision with the (left preceding, preceding, right preceding, left following, and right following) vehicles, lane with the highest attraction score and lane with the highest frontal gap. These features are extracted in numerical format from the HighD \cite{highDdataset} and \ac{CRASH} datasets. The numerical features are converted into linguistic categories based on literature and the data distribution along the lane-changing categories. For instance, the y-velocity is classified into \textit{leftMotion}, \textit{straightMotion}, and \textit{rightMotion} Similarly, the y-acceleration is classified into \textit{leftAcceleration}, \textit{zeroLateralAcceleration}, and \textit{rightAcceleration}. These thresholds are determined by analyzing the normal distribution of each feature category, following the methodology outlined in \cite{manzour2024vehicle}. The lane position is already a categorical feature.
For the distance from the center of the lane and time headway, thresholds are determined based on the normal distribution following the same approach as in the lateral velocity and acceleration. Thresholds for the \ac{TTC} are established based on previous studies \cite{saffarzadeh2013general}\cite{ramezani2020comparing}. A \ac{TTC} between zero and four seconds is considered high risk, between four and ten seconds as medium risk, and greater than ten seconds or negative values as low risk. For the lane attraction and frontal gap, thresholds are not applied. Instead, the lane with the highest attraction score is identified directly. Similarly, the lane with the highest numerical value for the frontal gap is considered. \Cref{tab:KG_classes} introduces the ontology of our \ac{KG} which provides a formal and general representation of the entities and their relationships within a \ac{KG}. Ontologies are essential in \acp{KG} because they serve as a schema for constructing the graph, ensuring consistency and improving its interpretability.

\begin{table*}[ht]
\caption{Ontology table which includes the definition of all entities (classes), their instances, and possible relations that can be connected to them.}
\vspace{-3mm}
\centering
\setlength{\tabcolsep}{5.5pt}
\label{tab:KG_classes}
\begin{tabular}{|c|c|c|c|}
\hline
Class               & Class Description                              & Instance                                  & Possible Relation         \\ \hline
intention           & Lane change intention of the vehicle           & LLC \textbar\ LK \textbar\ RLC                            & INTENTION\_IS             \\ \hline
lateralVelocity     & Lateral velocity                               & \{left,straight,right\}Motion & LATERAL\_VELOCITY\_IS     \\ \hline
                    &                                                & \{left,right\}Acceleration                          &                           \\
lateralAcceleration & Lateral acceleration                           & zeroLateralAcceleration                   & LATERAL\_ACCELERATION\_IS \\  \hline
                    &                                                & \{left,right\}LaneInTwoLanesRoad                    &                           \\
laneID              & Vehicle's lane position                        & \{left,center,right\}LaneInThreeLanesRoad                  & LANE\_ID\_IS              \\ \hline
                    &                                                & near\{left,right\}Marking                           &                           \\
laneMarkingOffset   & Lateral distance from lane marking             & centerOfTheLane                           & LANE\_MARKING\_STATE\_IS  \\ \hline
timeHeadway         & Time headway                                   & safe \textbar\ risky \textbar\ collision                  & TIME\_HEADWAY\_IS         \\ \hline

ttcLeftPreceding    & TTC with the left preceding vehicle            & \{low,medium,high\}RiskLeftPreceding                   & LEFT\_PRECEDING\_TTC\_IS  \\ \hline

ttcPreceding        & TTC with the preceding vehicle                 & \{low,medium,high\}RiskPreceding                       & PRECEDING\_TTC\_IS        \\ \hline

ttcRightPreceding & TTC with the right preceding vehicle & \{low,medium,high\}RiskRightPreceding & RIGHT\_PRECEDING\_TTC\_IS \\ \hline

ttcLeftFollowing    & TTC with the left following vehicle            & \{low,medium,high\}RiskLeftFollowing                   & LEFT\_FOLLOWING\_TTC\_IS  \\ \hline

ttcRightFollowing & TTC with the right following vehicle  & \{low,medium,high\}RiskRightFollowing & RIGHT\_FOLLOWING\_TTC\_IS  \\ \hline

attractionScore     & Lane with the highest attraction score         & \{left,center,right\}LaneHighestAttraction               & ATTRACTION\_SCORE         \\ \hline

frontGap          & Lane with the largest front gap                & \{left,center,right\}LaneLargestFrontGap                 & FRONT\_GAP                \\ \hline
vehicleID           & Child vehicle ID which changes every frame     & vehicle ID number (e.g. '445')            & HAS\_CHILD                \\ \hline
vehicle             & Generic entity pointing to every child vehicle & --                                        & Any                       \\ \hline
\end{tabular}
\vspace{-6mm}
\end{table*}
\subsection{Knowledge Graph Embedding Phase}
\begin{figure}[ht]
\centering
\includegraphics[width=\columnwidth]{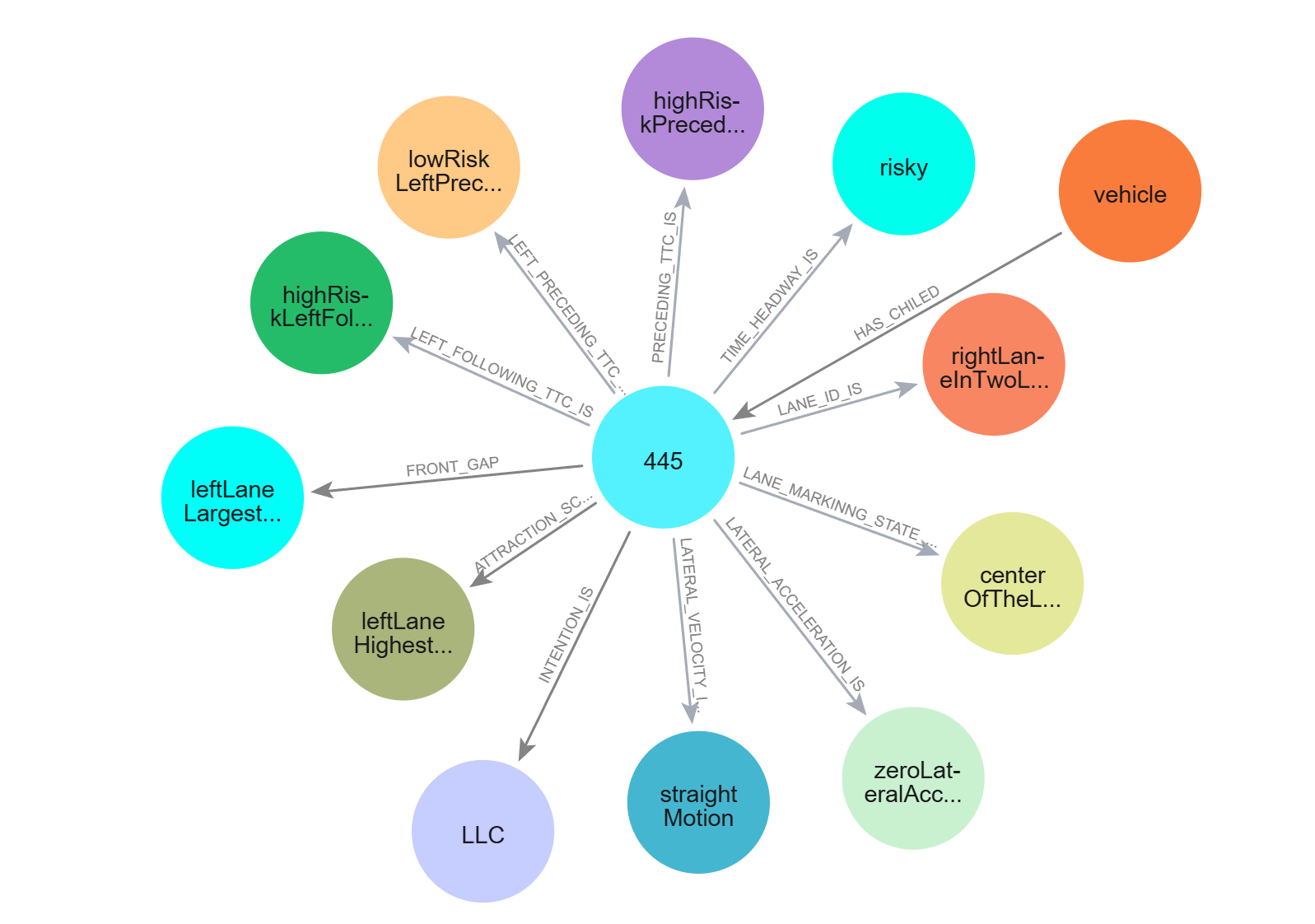}
\vspace{-7mm}
\caption{A KG instance for a single vehicle at one frame within the dataset.}
\label{fig:445}
\end{figure}

\begin{figure}[ht]
\centering
\includegraphics[width=\columnwidth]{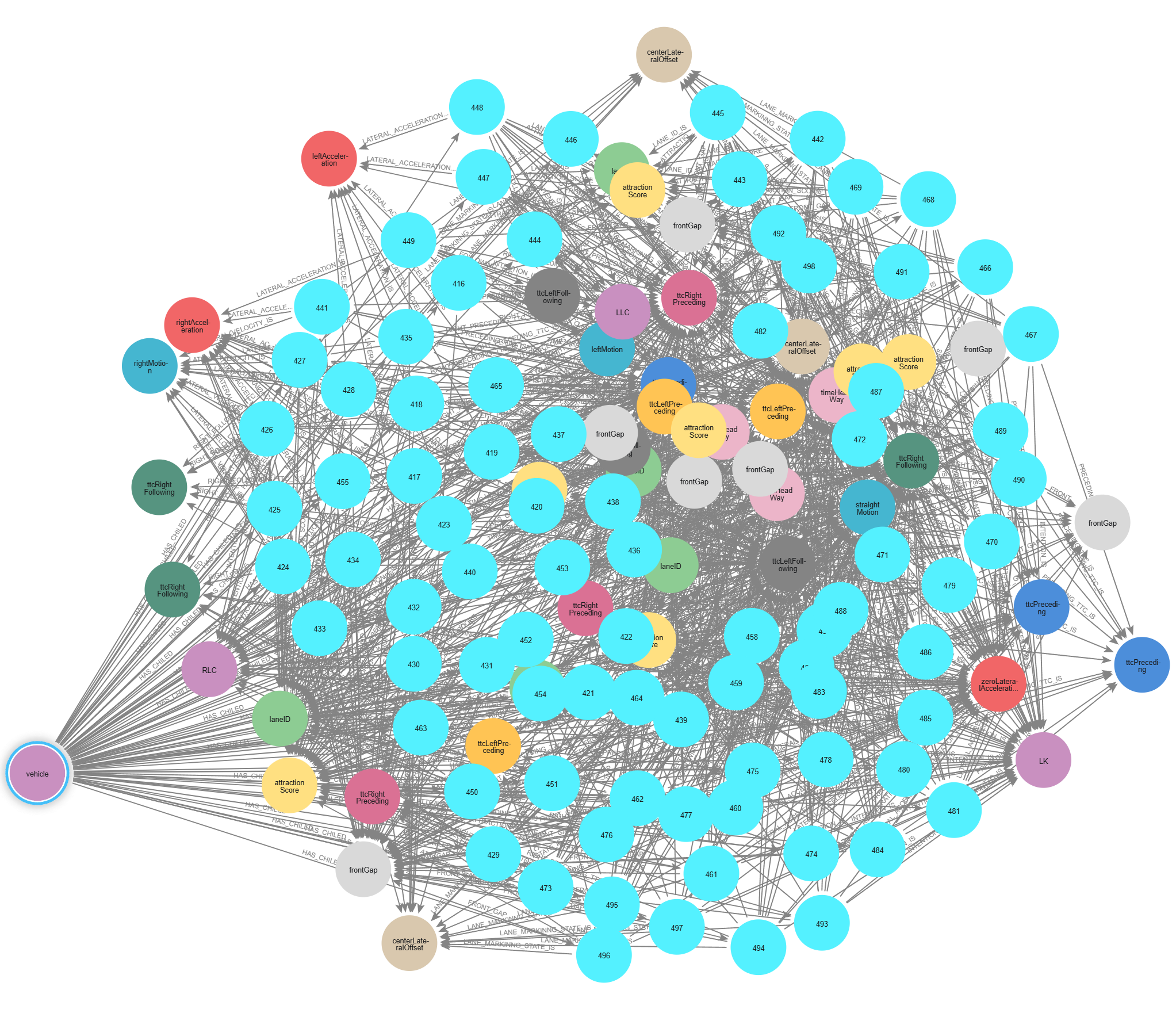}
\vspace{-7mm}
\caption{A sample KG representation with 80 instances from a total of $40K$ instances.}
\label{fig:90_instance_LQ}
\vspace{-6mm}
\end{figure}

A \ac{KG} is a directed heterogeneous multigraph, where nodes can have different types, and each pair of nodes can be connected by multiple types of relationships.
\acp{KG} are used to represent the data in the form of triples inside a CSV file. Each feature was expressed in one triple: subject, relation, and object. For example, the triple \textit{$<$vehicle 445, LATERAL\_ACCELERATION\_IS, leftAcceleration$>$} is mapped to \textit{$<$subject, relation, object$>$}. There is a generic entity called \textit{vehicle} which has several children such as \textit{vehicle 445}, \textit{vehicle 446}. Each vehicle ID is connected to its corresponding features. \Cref{fig:445} shows \ac{KG} instant for one vehicle from the full \ac{KG}. \Cref{fig:90_instance_LQ} presents KG instances for $80$ vehicles, given that the total number of instances used for training is $40K$, making it hard to fit in a single figure. It is important to note that in our \ac{KG} representation, \textit{vehicle 445} and  \textit{vehicle 446} are considered different vehicles, although they could represent the same vehicle in a real scenario. So, each vehicle will have a new child ID in each frame even if it is the same vehicle.
After that, we start to train/embed this knowledge graph. \ac{KGE} is a supervised machine learning technique that learns to represent (embed) the knowledge graph entities and relations into a low-dimensional vector space while preserving semantic meaning \cite{zheng2020dgl}. In this stage, we use the Ampligraph library \cite{ampligraph}. The scoring function used is the TransE function. The embedding size is set to $100$. The model parameters include using the Adam optimizer with a learning rate of $0.0005$, a batch size of $10,000$, and employing SelfAdversarialLoss. Additionally, for each positive triple in our dataset, we generate five negative triples, maintaining a corruption ratio of $5:1$. An early stopping criterion is used to monitor the \ac{MRR} metric during validation with a patience threshold of five validation epochs.
\subsection{Bayesian Inference and Prediction Phase}
\begin{figure*}[ht]
\centering
\includegraphics[width=\linewidth]{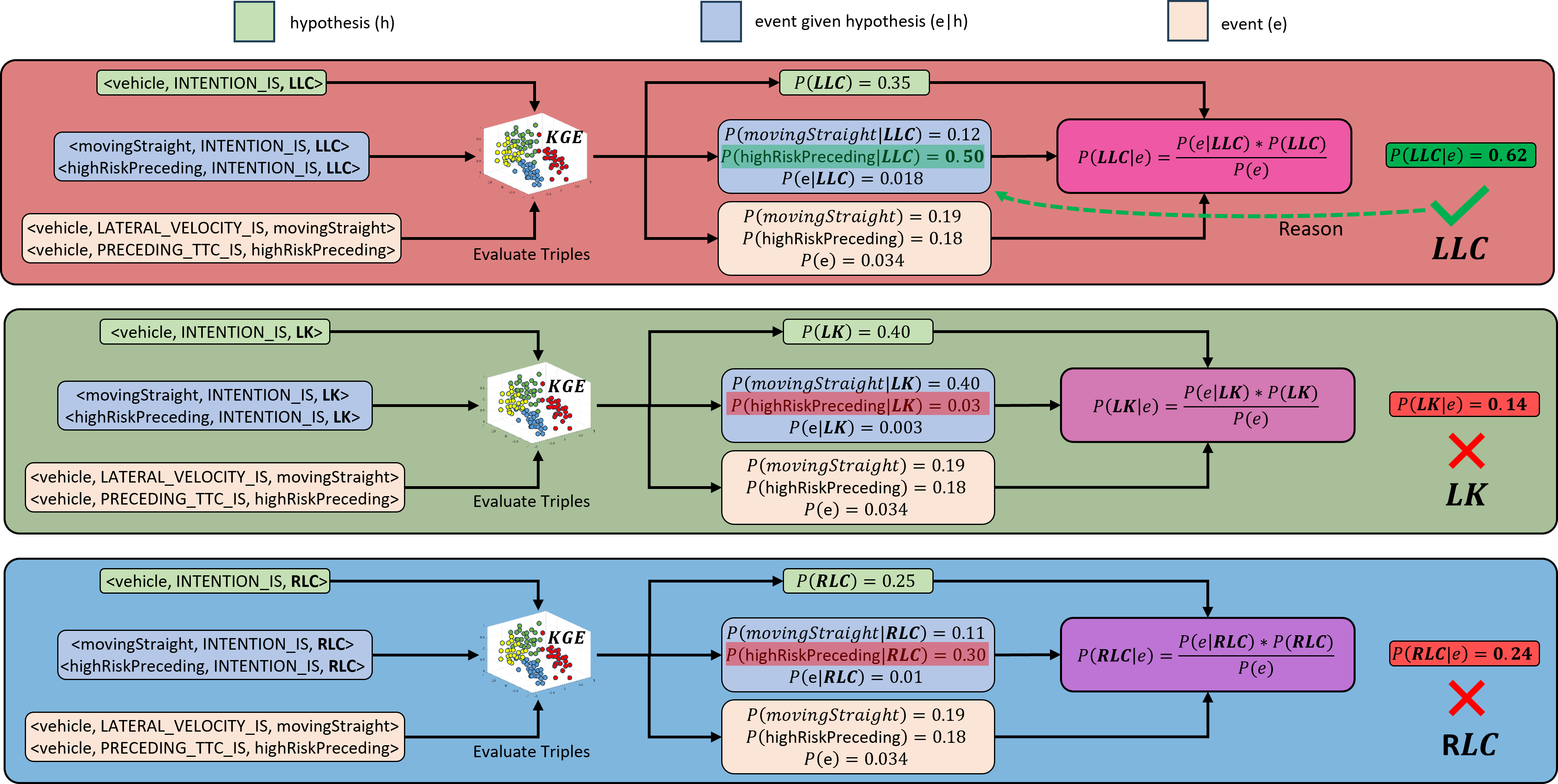}
\caption{Example of Bayesian inference reasoning for a left lane change prediction that shows the model traceability and transparency. The target vehicle will do the maneuver because $P(highRiskPreceding|LLC)$ is higher than $P(highRiskPreceding|LK)$ and $P(highRiskPreceding|RLC)$.}
\label{fig:bayesian_inference}
\vspace{-5mm}
\end{figure*}

In this stage, Bayesian inference is performed as a downstream task using the learned embeddings from the previous stage. This stage utilizes the Bayesian inference \cref{eq:bayesian}, which involves calculating the probabilities of a hypothesis \textit{(h)}, an event given the hypothesis \textit{(e\textbar h)}, and an event \textit{(e)}, to determine the probability of the hypothesis given the event, as shown in \cref{fig:bayesian_inference}.
Here, events are data obtained from sensors. After converting the data to the linguistic format, we formulate the triples related to these events and propose a hypothesis, for instance, the vehicle will make a left lane change. We then calculate \textit{P(LLC)} and \textit{(e\textbar LLC)} as indicated in \cref{fig:bayesian_inference}. For example, we ask the model to calculate the probability of a vehicle moving straight, given that it intends to make a left lane change. In case of the presence of multiple events, \cref{eq:evidence} and \cref{eq:evidence-hypothesis} are applied to determine the final values of \textit{P(e)} and \textit{P(e\textbar h)}, respectively.
\begin{equation}
\begin{aligned}
    P(h|e)=\frac{P(h)P(e|h)}{P(e)}
\end{aligned}
    \label{eq:bayesian}
\end{equation}
\begin{equation}
    P(e) = P(e_1) \times \dots \times P(e_n)
    \label{eq:evidence}
\end{equation}
\begin{equation}
     P(e|h) = P(e_1,\dots,e_n|h) = P(e_1|h) \times \dots \times P(e_n|h)
    \label{eq:evidence-hypothesis}
\end{equation}
These probabilities are evaluated using the embeddings obtained in the previous stage and are then fed into the Bayesian inference equation. This allows us to calculate \textit{P(LLC\textbar e)} which is the probability of the hypothesis (left lane change) given the events. However, this is just one hypothesis, we have to consider other possible maneuvers, which are lane keep or a right lane change. So, the probability for each maneuver is calculated, and the hypothesis with the highest probability is considered as the model's final prediction. Once the final prediction is obtained, its reasoning can be traced by comparing \textit{P(e\textbar h)} for each event across all hypotheses. As shown in the figure, \textit{P(highRiskPreceding\textbar LLC)} is higher than \textit{P(highRiskPreceding\textbar LK)} and \textit{P(highRiskPreceding\textbar RLC)}. However, this trend does not hold for the \textit{movingStraight} event. This indicates that the \textit{highRiskPreceding} event is the primary factor influencing the prediction in this situation and highlights the transparency of the model.

\subsection{RAG Explanation Phase}
After generating the prediction, we need our architecture to justify why our model provided this prediction and reason about the target vehicle decision making in natural language using an \ac{LLM}. However, the challenge is that \acp{LLM} have access to a wide public database, which does not fit our purposes. We require the justifications to be based only on the provided \acp{KG} derived from the HighD and the \ac{CRASH} datasets, which are private databases, not public ones. We also need the justifications to be more reliable and verifiable. This can be achieved by grounding them in external, independently verifiable knowledge sources rather than relying solely on internal knowledge acquired during the training process. \ac{RAG} is used to address this issue. First, we organize our \ac{KG} triples into chunks, with each chunk representing one sample from our data. We then embed these chunks using an embedding model and store the resulting embeddings in a vector database.  \textit{all-MiniLM-L6-v2} from Hugging Face is the used embedding model and \textit{Chroma} is the used vector database. When a user submits a query, the system embeds the query and retrieves chunks with embeddings that are most similar to the query. These retrieved chunks are then augmented to the query, and to a prompt that guides our model to the type of needed responses To ensure that the responses are accurate and based solely on the private \ac{KG} data. The final augmented query is then fed into an \textit{OpenAI GPT-4} \ac{LLM} to generate the required response.
\section{Results}\label{sec:results}
In this section, we present the results of our work, which aims to predict both safe and risky lane changes using a unified model and consistent labels. We compare our approach with existing studies that focus solely on predicting safe lane-changing maneuvers. Furthermore, we validate our model by integrating it into an IDM within the CARLA simulation environment to test the IDM reactions with and without our prediction model. Finally, we demonstrate the explainability of our approach by utilizing the \ac{RAG} module to provide natural language explanations for the model's predictions.

\subsection{Model Performance and Comparative Analysis}
Regarding safe lane-changing maneuvers, our model achieved an average f1-score of $90\%$ over the interval [0,4] seconds. We identified four relevant articles addressing safe lane-changing maneuvers using the HighD dataset, and compared their results with ours in \cref{tab:comparison_1}.
\begin{table}[ht]
\caption{Performance comparison of models predicting \underline{\textbf{safe}} lane changes on the HighD dataset using the f1-score metric.}
\vspace{-4mm}
\begin{center}
\setlength{\tabcolsep}{4pt}
\begin{tabular}{|c|c|c|c|c|c|c|c|c|c|}
\hline
Pred. time (s) & 0.5s & 1s & 1.5s & 2s & 2.5s & 3s & 3.5s & 4s\\
\hline
$\cite{xue2022integrated}$   & 98.2  & 97.1  & 96.6 & 95.2 & -- & -- & -- & --\\
\hline
$\cite{gao2023dual}$   & \underline{\textbf{99.2}}  & \underline{\textbf{99.0}}  & 97.6 & 91.8 & -- & -- & -- & --\\
\hline
$\cite{manzour2024vehicle}$  & 97.7  & 97.9  & 98.1 & \underline{\textbf{98.0}} & \underline{\textbf{97.2}} & \underline{\textbf{93.6}} & \underline{\textbf{82.8}} & \underline{\textbf{66.5}}\\
\hline
Ours  & 95.0 & 95.5 & 95.5 & 95.0 & 94.0 & 90.0 & 80.5 & 65.5\\
\hline
\hline
Interval (s) & [0,1] & (1,2] & (2,3] & (3,4] & [0,4] & -- & -- & --\\
\hline
$\cite{peng2024lc}$   & \textbf{98.5}  & \textbf{98.9}  & \textbf{98.1} & \textbf{93.0} & \textbf{97.1} & -- & -- & --\\
\hline
Ours & 95.5 & 95.8  & 92.4 & 76.0 & 90.0 & -- & -- & --\\
\hline
\end{tabular}
\label{tab:comparison_1}
\end{center}
\vspace{-3mm}
\end{table}
The first two articles (\cite{xue2022integrated,gao2023dual}) report results at specific times. Their models exhibit higher performance than ours at time points (0.5 to 2 seconds). Also, no results are indicated for longer horizons. Additionally, their models employ deep learning techniques, such as \acp{LSTM} and transformers, which are considered "black-boxes" and lack transparency and interpretability. In contrast, our model is interpretable and provides transparency to users, as indicated in \cref{fig:bayesian_inference}. The third article \cite{manzour2024vehicle} also demonstrates consistently high-performance scores. However, like the previous two models, this model was trained only to predict safe lane changes, whereas our model is designed to predict both safe and risky lane changes. Another study reports remarkable results for safe lane-changing maneuvers on the HighD dataset, presenting performance over intervals from [0,1], (1,2], (2,3], (3,4], and [0,4] seconds. Their model utilizes an \ac{LLM} with a chain-of-thought approach. While the model's results are impressive, the model is based on numerical inputs and lacks transparency and verifiability, similar to other "black-box" models where the internal workings of the model are not accessible, making it difficult to understand which inputs influence the predictions. In contrast, our model provides high transparency as mentioned previously and it also deals with
risky situations (as mentioned in \cref{tab:SOTA}) which are not considered in state-of-the-art systems. For risky lane changes, results are shown in \cref{tab:risky_LC_results}.
\vspace{-3mm}
\begin{table}[ht]
\caption{Results of our model on \underline{\textbf{risky and safe}} lane changes based on the f1-score metric. Model $1$ is the original model trained on both CRASH and HighD datasets. Model $2$ is trained on HighD only.}
\vspace{-4mm}
\begin{center}
\begin{tabular}{|c|c|c|c|c|c|}
\hline
Pred. time (s) & [0,1]  & (1,2]  & (2,3]  & (3,4] & [0,4]\\
\hline
\makecell{Model $1$ tested on\\ CRASH (risky)}  & 99.3  & 91.9 & 91.5 & 81.8 & 91.5\\
\hline
\makecell{Model $1$ tested on\\(CRASH+HighD) (mixed)} & 95.7  & 94.9 & 91.7 & 75.7 & 89.4\\
\hline
 \makecell{Model $2$ tested on CARLA} & 89.3 & 36.1 & 30.6 & 21.9 & 50.7\\
\hline
\end{tabular}
\label{tab:risky_LC_results}
\end{center}
\vspace{-3mm}
\end{table}
The model attained an f1-score of $91.5\%$ over the interval [0,4] seconds. To the best of our knowledge, no prior studies have addressed these specific types of maneuvers. Therefore, we present our results accordingly without comparison to other studies. To demonstrate the significance of the risky dataset, we trained another model (model $2$) only on the HighD dataset and tested it on the risky lane changes included in the CRASH dataset. The model is unable to predict risky lane changes effectively as shown in the last row of \cref{tab:risky_LC_results}, highlighting the importance of including a dataset that contains risky lane changes in the training process.
\subsection{Validation in Simulation and Real-World Scenarios}
\begin{figure}[ht]
\centering
\includegraphics[width=\columnwidth]{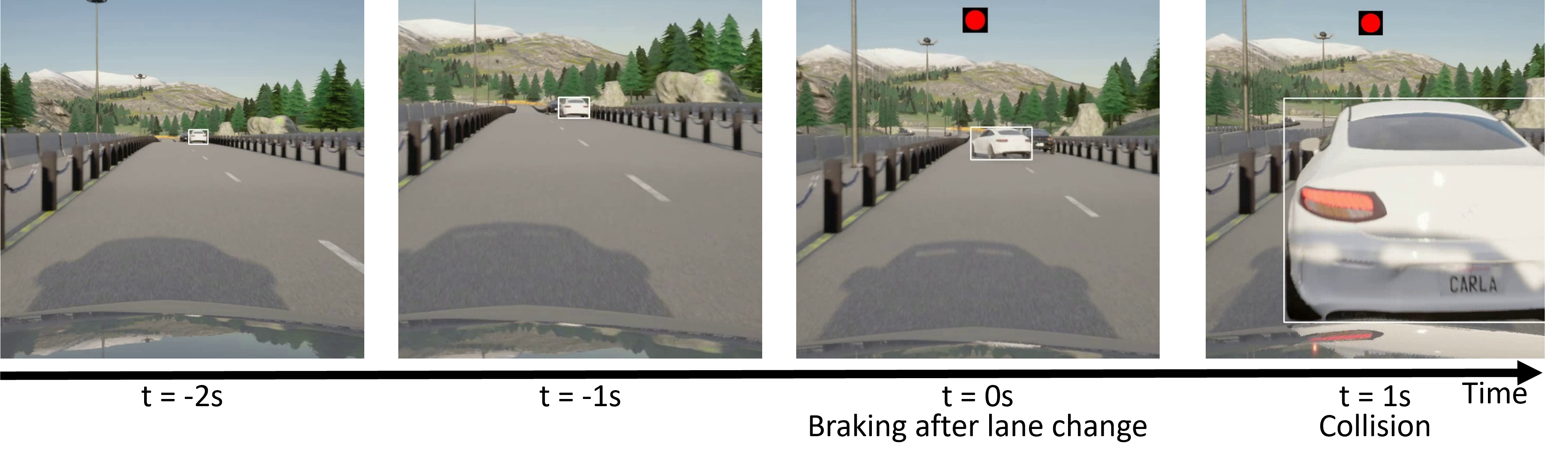}
\vspace{-8mm}
\caption{The ego vehicle is moving forward safely (first two images). Then, the target vehicle unexpectedly changed lanes to the left to be in the ego vehicle's lane to avoid a slower preceding vehicle (third image). This leads to a collision due to insufficient braking time (fourth image).}
\label{fig:IDM_no_prediction}
\vspace{-2mm}
\end{figure}

\begin{figure}[ht]
\centering
\includegraphics[width=\columnwidth]{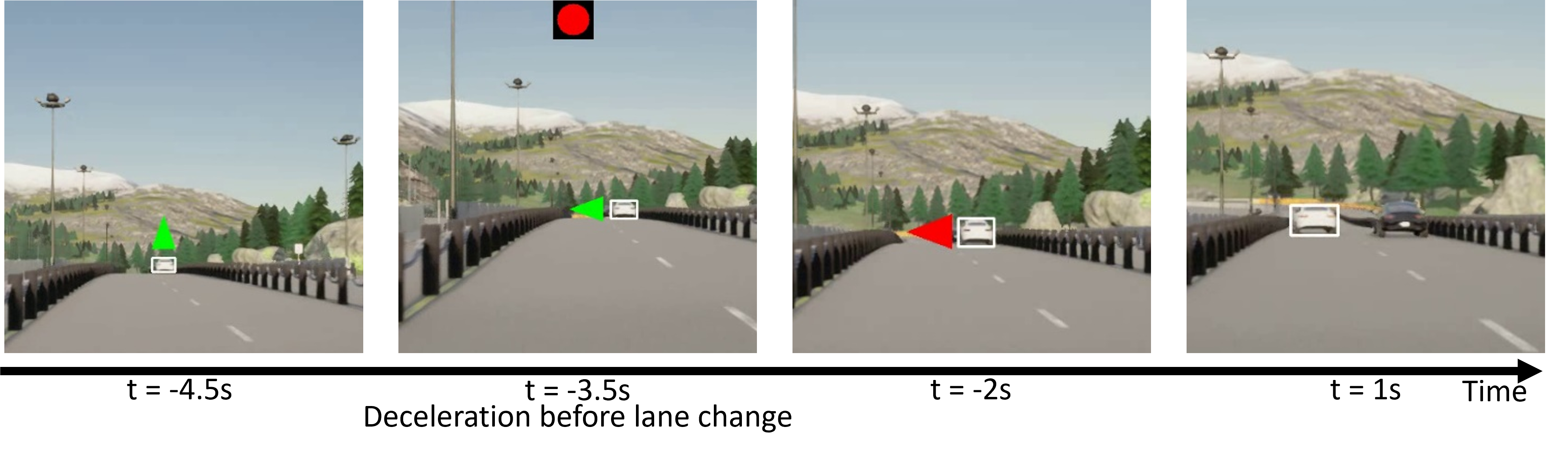}
\vspace{-8mm}
\caption{In the first image, the ego vehicle is moving forward normally and predicts a lane-keep behavior for the target vehicle. Then, the target vehicle is predicted to make a safe lane change to the left, at that moment the ego vehicle started to decelerate for more safety (second image). The third image shows the anticipation of a risky lane change. Finally, both vehicles continue to move safely without collision (fourth image).}
\label{fig:IDM_with_prediction}
\vspace{-3mm}
\end{figure}

\begin{figure}[ht]
\centering
\includegraphics[width=\columnwidth]{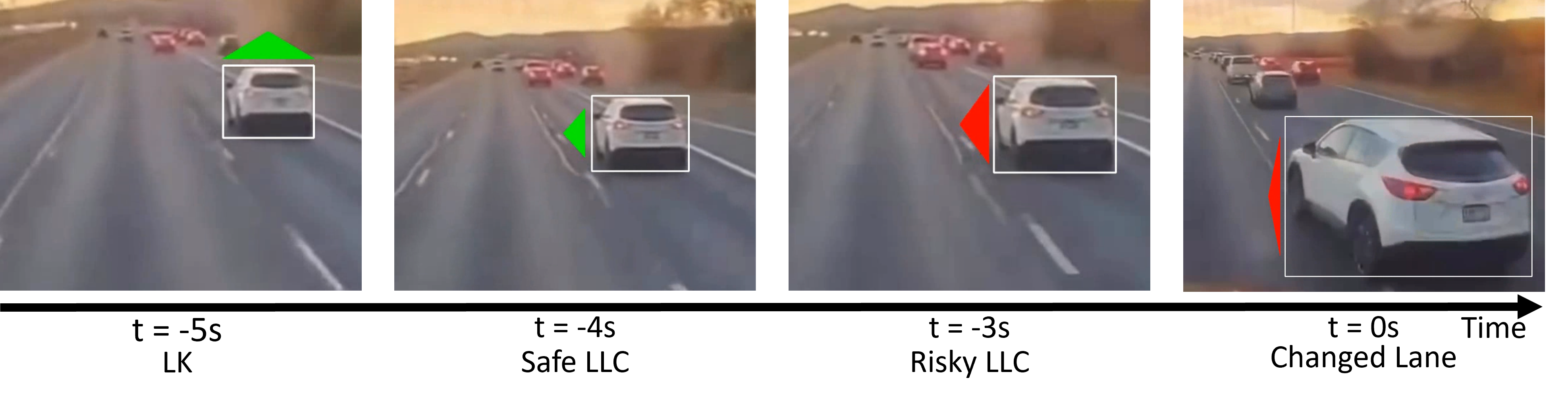}
\vspace{-8mm}
\caption{The model anticipates a risky left lane change three seconds in advance: In the first image, the white target vehicle is moving straight as expected. In the second image, the model predicts a safe left lane change. However, the target vehicle continues moving forward, resulting in a risky left lane change shown in the third image, leading to a collision with the ego vehicle in the fourth image. If the ego vehicle had reacted based on the model’s early predictions, this collision could have been avoided.}
\label{fig:real_scenario_prediction}
\vspace{-2mm}
\end{figure}

Initially, some scenarios are validated using the \ac{IDM} integrated on the ego vehicle in CARLA without integrating our prediction model. In these cases, the IDM didn't have enough time to brake, ending up with a collision. For example, \cref{fig:IDM_no_prediction} shows a scenario where the ego vehicle moves forward and the target vehicle is on the right front side. The \ac{IDM} detects no obstacles ahead and increases its velocity. However, the target vehicle faces a risk with the vehicle in front of it and needs to change lanes. In this situation, the \ac{IDM} does not know the intention of this vehicle or that the target vehicle is at risk which will lead to the execution of a risky left lane change. Consequently, the ego vehicle suddenly encounters another vehicle changing lanes directly ahead, leaving insufficient time to brake.
After integrating the prediction model, it anticipates that the target vehicle will execute a risky/sudden left lane change, as shown in \cref{fig:IDM_with_prediction}. This anticipation allowed the ego vehicle to decelerate smoothly, rather than braking abruptly. The model provides the ego vehicle with awareness of the upcoming risky lane change, enabling it to decelerate in anticipation of the target vehicle's maneuver. Eventually, the target vehicle changed lanes to the left, while the ego vehicle continued moving safely.
Then, our model is tested on real-world videos that contain risky lane changes. The purpose of these scenarios is to demonstrate that our model functions effectively not only with simulation data but also in real-world situations. In \cref{fig:real_scenario_prediction}, we have some captures from a real-world video where our model anticipated the risky left lane change in advance. By integrating our model into the vehicle, we can potentially avoid near-crashes or accidents resulting from risky lane changes and understand the target vehicle's decisions.
Finally, \autoref{tab:media} contains two hyperlinks. The first one directs to a website to access the \ac{CRASH} dataset, and shows an interactive graph for a sample from our designed \ac{KG}. The second hyperlink directs to a YouTube playlist that contains some multimedia videos of different scenes including the scenes discussed in this section.
\vspace{-3mm}
\begin{table}[ht]
\caption{The first hyperlink leads to a website that includes a summary of the work, an interactive \ac{KG}, and access to the CRASH dataset. The second hyperlink is a YouTube playlist featuring additional video demonstrations.}
\vspace{-4mm}
\begin{center}
\begin{tabular}{|c|c|}
\hline
      Media          & Hyperlink\\
\hline
Website   & \href{https://butternut-clef-a48.notion.site/Explainable-Lane-Change-Prediction-for-Near-Crash-Scenarios-Using-Knowledge-Graph-Embeddings-and-Ret-162eb56a8de78064966cc55607d5b11a?pvs=4}{Website\_ExplainableRiskyLaneChangePrediction}\\
\hline
YouTube   & \href{https://www.youtube.com/playlist?list=PLAeK3AuwxenFaWQut8umkxfUXf6qoB_s7}{YouTube\_ExplainableRiskyLaneChangePrediction}\\
\hline
\end{tabular}
\label{tab:media}
\end{center}
\vspace{-6mm}
\end{table}

\subsection{Natural Language Explanation Through RAG}
\begin{figure}[ht]
\centering
\includegraphics[width=\columnwidth]{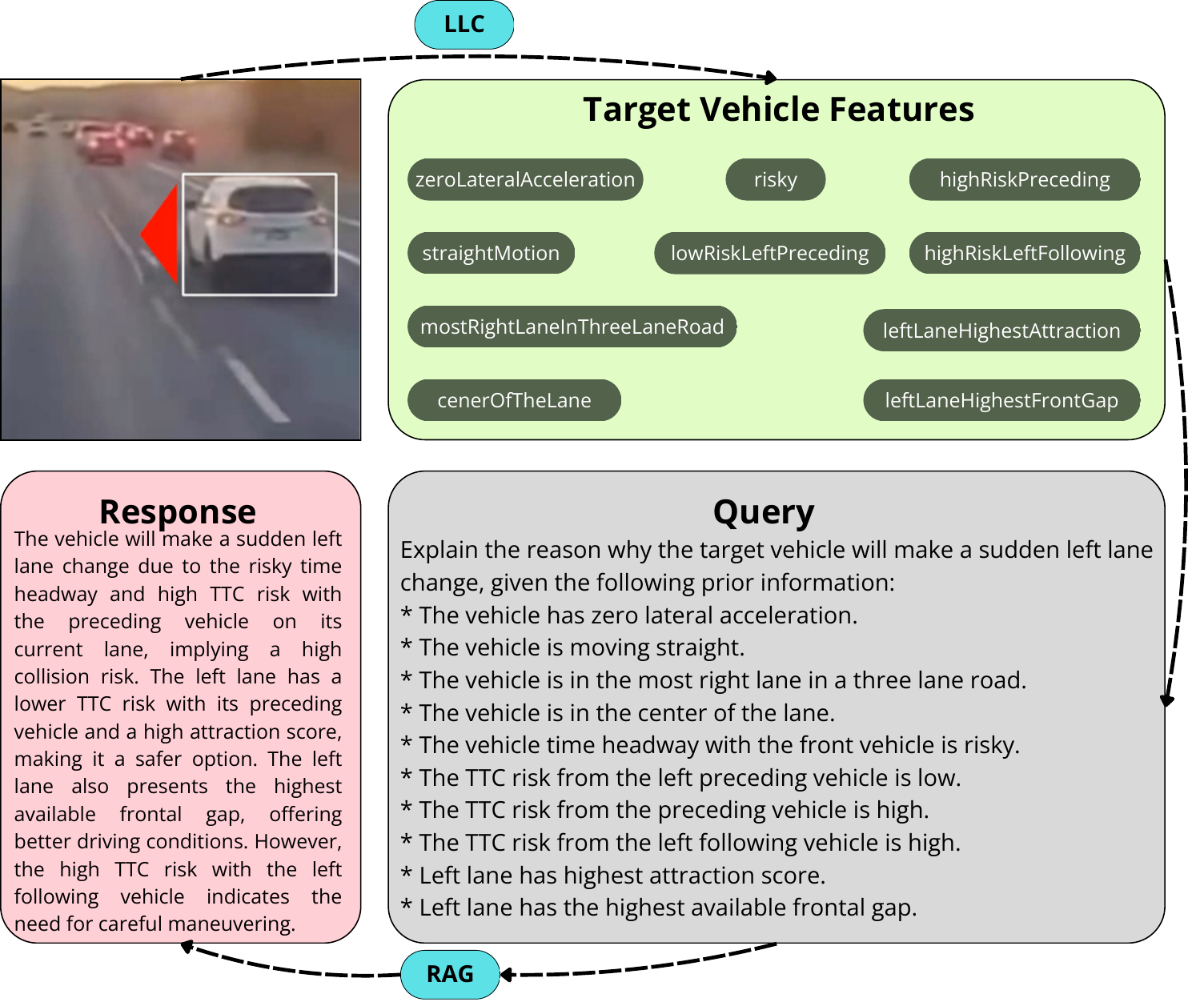}
\vspace{-6mm}
\caption{Query formation by merging the prediction with the linguistic inputs. Then, the query is fed to the \ac{RAG} model to provide the explanation of the predicted risky left lane change.}
\label{fig:RAG_result}
\vspace{-5mm}
\end{figure}

After validation of the model, we analyze one of the captions extracted from the same real-world scenario, then we feed the linguistic inputs obtained from this capture along with the prediction to the \ac{RAG} model. \cref{fig:RAG_result} shows the formed query obtained from the linguistic inputs and the predicted maneuver. Then, the query is fed to the \ac{RAG} model which outputs an explanation that illustrates the model's ability to provide clear, reasonable, and precise natural language explanations for the target vehicle's predicted maneuver.
\section{Conclusion}\label{sec:concliusions}
In this work, we anticipated safe and risky lane-changing maneuvers using \ac{KGE} and Bayesian inference. We trained our model on the HighD dataset for safe lane changes and on our own created \ac{CRASH} dataset for risky lane changes. Our model achieved an f1-score of $90\%$ for predicting safe lane changes with an anticipation interval of $4$ seconds and an f1-score of $91.5\%$ for risky lane changes within the same anticipation window. In terms of numerical performance, our model is competitive with other state-of-the-art systems in anticipating safe lane changes and contributes by anticipating risky lane changes. Moreover, the model's performance goes beyond numerical results by enhancing transparency and interpretability of the prediction process. We validated our model by integrating it to an \ac{IDM} in the CARLA simulator, and it is proven that predicting risky lane changes provides the ego vehicle with more time to react and avoid near-crash situations. Additionally, we employed \ac{RAG} to enhance the model's explainability by providing natural language explanations for its predictions. Future work can focus on the decision-making process of the ego vehicle by incorporating advanced actions based on the predicted lane change of the target vehicle.
\vspace{-2mm}
\section*{Acknowledgment}
This research has been funded by the HEIDI project of the European Commission under Grant Agreement: 101069538.

\ifCLASSOPTIONcaptionsoff
  \newpage
\fi



%

\bibliographystyle{IEEEtran}
\bibliography{IEEEabrv,bibtex/bib/IEEEexample}

%








\end{document}